\documentclass[a4paper, 10pt, conference]{ieeeconf}      
\usepackage{FG2026}

\FGfinalcopy 

\usepackage{caption}
\usepackage{multicol}
\usepackage{graphicx}
\usepackage{amsmath}
\usepackage{amssymb}
\usepackage{booktabs}
\usepackage{bm}
\usepackage{bbm}
\usepackage{color}
\usepackage{colortbl}
\usepackage{xcolor}
\usepackage{listings}
\usepackage{url}
\usepackage[symbol]{footmisc}
\usepackage{pifont}
\definecolor{Gray}{gray}{0.9}
\definecolor{LightCyan}{rgb}{0.88,0.95,1}
\usepackage{todonotes}
\usepackage{tabularx}
\usepackage{verbatim}
\usepackage{cuted}
\usepackage{arydshln}
\usepackage{multirow}

\newcommand{\rev}[1]{\textcolor{blue}{#1}}
\renewcommand{\emph}[1]{\textit{#1}}
\newcommand{\etal}{\emph{et al}.}

\IEEEoverridecommandlockouts                              
\title{\LARGE \bf
ReactionMamba: Generating Short \& Long Human Reaction Sequences
}

\author{\parbox{16cm}{\centering
    {\large Hajra Anwar Beg$^{1,2}$, Baptiste Chopin$^3$, Hao Tang$^4$, Mohamed Daoudi$^{1,2}$}\\
    {\normalsize
    $^1$ Univ. Lille, CNRS, Centrale Lille, Institut Mines-Télécom, UMR 9189 CRIStAL, F-59000 Lille, France\\
    $^2$ IMT Nord Europe, Institut Mines-Télécom, Univ. Lille, Centre for Digital Systems, F-59000 Lille, France\\
    $^3$ da/sec – Biometrics and Security Research Group, Hochschule Darmstadt, Germany\\
    $^4$ Peking University}}
}

\begin{document}

\ifFGfinal
\thispagestyle{empty}
\pagestyle{empty}
\else
\author{Anonymous FG2026 submission\\ Paper ID \FGPaperID \\}
\pagestyle{plain}
\fi
\maketitle

\newcommand{\teaserCaption}{\textbf{Examples of reactive 3D motion generated with our proposed model for the NTU120-AS dataset Cheers and drink class (top) and the ReMoCap Ninjutsu dataset (bottom)}. Here we synthesize the 3D pose of the reactor (\textcolor{green}{green}) conditioned on the 3D motion of the actor (\textcolor{blue}{blue}) and the initial pose of the reaction.}

\definecolor{gtbody}{RGB}{89,147,203}
\definecolor{predbody}{RGB}{101,167,158}

\begin{strip}
    \centering
    \includegraphics[width=\linewidth]{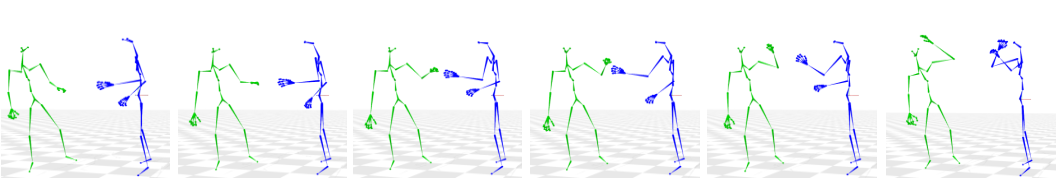}
    \captionof{figure}{\textbf{Examples of reactive 3D motion generated with our proposed model for the NTU120-AS dataset Cheers and drink.} Here we synthesize the 3D pose of the reactor (\textcolor{green}{green}) conditioned on the 3D motion of the actor (\textcolor{blue}{blue}) and the initial pose of the reaction.}
    \label{fig:teaser}
\end{strip}
\vspace{-10mm}

\begin{abstract}
We present ReactionMamba, a novel framework for generating long 3D human reaction motions. ReactionMamba integrates a motion VAE for efficient motion encoding with Mamba-based SSMs to decode temporally consistent reactions. This design enables ReactionMamba to generate both short sequences of simple motions and long sequences of complex motions, such as dance and martial arts. We evaluate ReactionMamba on three datasets—NTU120-AS, Lindy Hop, and InterX—and demonstrate competitive performance in terms of realism, diversity, and long-sequence generation compared to previous methods, including InterFormer, ReMoS, Ready-to-React, and Think-Then-React while achieving substantial improvements in inference speed. The code of the paper will be available if the paper is accepted.
\end{abstract}
\section{Introduction}
In recent years, there has been impressive progress in human motion generation under various user-specified conditions, such as music pieces~\cite{MusicToHumanMotionGenerationCVPR2023} and text prompts~\cite{petrovich23tmr}. However, most of these approaches focus on a single subject setting~\cite{chen2023executing, petrovich22temos} and, as a result, neglect one essential aspect of human motion: rich human-to-human interactions. One important aspect of human interaction is that the action of one person will usually elicit a reaction from another person. Synthesizing human-to-human interactions introduces advanced capabilities for character animation tools and software, with applications across commercial and entertainment media~\cite{Hanser2010}, interactive mixed and augmented reality~\cite{ArjanVisComput2007}, social robotics~\cite{ICRA2019}, and avatar interaction~\cite{ChopinICIAP2023}. 

This led recent research to take interest in the task and propose various architectures to generate reaction motions~\cite{ghosh2024remos, ChopinInterformer2023, surveyhumaninteractionmotion2025}. However, human interaction reactions present a more challenging task due to the motion reactions, which are highly complex and have low periodicity. They require extremely accurate modeling of both spatial and temporal features, without which the generated motion looks unrealistic. While current state-of-the-art methods ~\cite{ghosh2024remos, ChopinInterformer2023} can now generate reactive motion for simple and short actions they still struggle with longer sequences. Indeed short sequences often include simple motion such as boxing or kicking that can be more easily modeled due to a direct body contact leading to a clear reaction. Longer sequences however often include more complex interactions, such as dancing, where the relation between the action and the reaction is not always as direct and obvious. Modeling longer interaction sequences requires specific architectures, different from those focused on short sequences, to achieve high quality reaction generation. Precursor works on reaction generation have represented the task as a translation problem which directed them towards the Transformer ~\cite{ChopinInterformer2023}. However, this led to low diversity generation and limited performance in the long term. 
 The self-attention mechanism ~\cite{vaswani2017attention} computes interactions between every pair of tokens in the input sequence. This leads to quadratic time complexity in relation to the sequence length, which becomes a bottleneck for processing long sequences. As a result, it is essential to explore a new architectural paradigm that can handle long-range dependencies while maintaining linear computational complexity to support motion generation tasks.
 Recent advances have led to renewed interest in State-Space Models (SSMs)~\cite{NEURIPS2021SSM}. SSMs are a class of mathematical models used for modeling dynamic systems over time, often in fields like control theory, signal processing, and time series analysis. They have several advantages, particularly in their ability to model temporal dependencies for long sequences~\cite{NEURIPS2021SSM}. The efficiency in  managing long sequences primarily due to its implementation of convolutional computations, which efficiently process sequential data~\cite{gu2022efficiently}. Convolutional operations reduce the complexity associated with handling long-range dependencies by applying local filters to the sequence such as Mamba~\cite{gu2024mambalineartimesequencemodeling}. In this paper, we propose ReactionMamba (Fig.~\ref{fig:overview}) a new architecture for reaction motion generation that matches the state-of-the-art while increasing inference speed by a hundredfold. Figure~\ref{fig:teaser} shows some examples results obtained by our framework on NTU120-AS~\cite{xu2024regennet}.

\noindent \textbf{Contributions.} Our contributions to the field of reaction motion generation can be summarized as :
\begin{itemize}[noitemsep,topsep=0pt]
    \item We present \emph{ReactionMamba}, a simple yet effective framework that leverages Mamba-based state space models to generate human motion reactions without relying on any auxiliary input or external information.
    \item \emph{ReactionMamba} is a Variational AutoEncoder (VAE)~\cite{VAE2014} made up of three interconnected modules, an encoder, a conditioning module, and a decoder. The SSM-based encoder encodes the reaction into a sequence of continuous latent variables in an efficient way, and a SSM-based decoder is then designed to capture information hidden in the latent representations of the pose, combined with the action motion sequence and the initial frame of the reaction and to reconstruct the reaction.
    \item Unlike conventional deterministic methods~\cite{julieta2017motion}, our method uses random latent sampling at inference, conditioned by the projected action sequence and the projected initial reaction pose, providing diverse and realistic reaction sequences.
    \item The \emph{ReactionMamba} framework demonstrates strong performance on a range of two-person action–reaction scenarios, including short-duration interaction categories from NTU120-AS \cite{xu2024regennet}, longer-duration categorized interactions from InterX [54], and more complex two-person interactions such as Lindy Hop dancing [20], while substantially improving inference speed.
\end{itemize}

\section{Related work}

\noindent \textbf{Single Subject Human Motion Synthesis}.
Single subject motion synthesis has seen remarkable progress in recent years. To cite some examples, conditional VAE~\cite{petrovich22temos, dabral2022mofusion} have been proposed to map textual descriptions to human motion. More recent approaches, such as Editable Dance GEneration (EDGE),~\cite{tseng2023edge} and Dancing to Music~\cite{lee2019dancing2music}, introduce frameworks for the generation of music-conditioned motion. However, these methods often require additional conditions to effectively model emotional expressiveness. In addition, these approaches focus on the generation of single-human while ReactMamba learns the synchronization between two interacting individuals directly from one person’s motion, without the need for additional prompts or labels. 


\noindent \textbf{Reactive Motion Generation}. Reactive motion generation focuses on generating human motion in response to external agents.
Reactive motion generation is inherently more challenging than generating a single human motion sequence. In~\cite{Yang2020}, a motion graph synthesis technique is proposed that generates gestures and body motions in a dyadic conversation. Huang \etal~\cite{HuangECCV2014} proposed to represent dual-agent interactions as an optimal control problem, where the actions of the initiating agent induce a cost topology over the space of reactive poses – a space in which the reactive agent plans an optimal pose trajectory. Chopin \etal ~\cite{ChopinInterformer2023} proposed \emph{InterFormer}, an interaction transformer with both spatial and temporal attention to generate reactive motions given some initial seed poses of both characters. It incorporates an interaction distance module that integrates softmax-scaled joint distances into the attention matrix. Ghosh \etal ~\cite{ghosh2024remos} introduced Reactive Motion Synthesis (\emph{ReMoS}), a diffusion-based diffusion framework incorporating spatio-temporal cross-attention and hand-interaction-aware cross-attention. 
These works primarily focused on generating high-quality short sequences spanning 2 to 5 seconds (around 50 frames), with \emph{InterFormer}~\cite{ChopinInterformer2023} noting an accumulation of errors in longer sequence generation.  
More recently, Cen \etal~\cite{ready_to_react_repo} proposed \emph{Ready-to-React}, an online reaction policy for two-character interaction generation, where each character independently predicts its next pose in a streaming manner using an auto-regressive model with a diffusion head, enabling long-term interaction generation without requiring future motion information. 
Tan \etal~\cite{tan2025think} proposed \emph{Think-Then-React}, an LLM-based framework for online and unconstrained action-to-reaction generation, which introduces an explicit thinking stage to infer action intent and periodically re-thinks during inference to mitigate semantic error accumulation. 
Although these attention-based approaches produce high-quality results, they face scalability challenges and high memory cost as the number of persons or frames increases, owing to the quadratic complexity of the attention mechanism ~\cite{vaswani2017attention}. In contrast, our method leverages \emph{Mamba}~\cite{mamba,mamba2}, which alleviates this limitation.

\noindent \textbf{State Space Models and Human Motion Synthesis}. Despite the well-known superiority of transformer-based architectures in vision tasks ~\cite{vaswani2017attention,dosovitskiy2020image,li2023multi,chopin2023interaction}, one of their most critical limitations lies in the quadratic complexity of the self-attention mechanism, which results in prohibitive memory consumption and computational overhead when dealing with long sequences or high-resolution visual inputs. This constraint has motivated an increasing body of research aiming to develop alternative sequence modeling paradigms that retain the representational power of transformers while significantly reducing their computational burden.

In this context, SSMs, inspired by classical control theory, have recently emerged as a powerful and efficient family of architectures for sequential data modeling~\cite{mamba,mamba2}. Unlike transformers, which rely on pairwise attention ~\cite{vaswani2017attention}, SSMs leverage parameterized state transitions to capture long-range dependencies in a linear-time manner. Mamba ~\cite{mamba}, for instance, introduces a selective SSM formulation that incorporates time-varying parameters into the state-space framework, together with a hardware-aware algorithm design, leading to substantial improvements in both training efficiency and inference scalability.

Thanks to these properties, SSMs have rapidly been adopted across a wide spectrum of visual tasks, ranging from video classification, comprehension, and segmentation ~\cite{islam2022long,nguyen2022s4nd,wang2023selective,yang2024vivim}, to movie scene detection ~\cite{islam2023efficient}. Their applications further extend to cross-modal and generative domains, including text-to-motion generation ~\cite{ZhangECCVMAMBA}, image compression ~\cite{zeng2025mambaic}, dance video synthesis ~\cite{tang2025spatial}, and image segmentation ~\cite{ma2024u,ruan2024vm,liu2024swin,xing2024segmamba}, demonstrating their versatility and growing impact in computer vision. 

Recently, Zeyu \etal~\cite{ZhangECCVMAMBA} proposed Motion Mamba, which achieves state-of-the-art performance in human motion synthesis by generating long, coherent motion sequences. Motion Mamba is a latent-diffusion text-to-motion model that replaces conventional Transformer-based temporal modeling with Mamba (selective state-space models) to effectively process long-range dependencies while maintaining inference efficiency. While Motion Mamba focuses on text-driven single-person motion generation, ReactionMamba addresses reaction synthesis, generating a reactor's 3D motion in response to an observed actor.

Tanke \etal~\cite{TankeCVPRW2025} introduced Dyadic Mamba for text-driven dyadic motion generation. Dyadic Mamba employs an SSM/Mamba-based diffusion model to generate text-conditioned, two-person interactions that remain coherent over extended durations. In contrast, ReactionMamba utilizes a VAE-Mamba framework to synthesize a reactor’s 3D motion conditioned on the actor’s movement and the reaction's initial pose, bypassing the need for text prompts. Furthermore, because Dyadic Mamba relies on iterative multi-step denoising, it entails significant computational costs during inference; ReactionMamba avoids this overhead through its one-step VAE-based formulation.




\section{The Proposed Method}

\begin{figure*}[!ht]
\centering 
\includegraphics[width=1\linewidth]{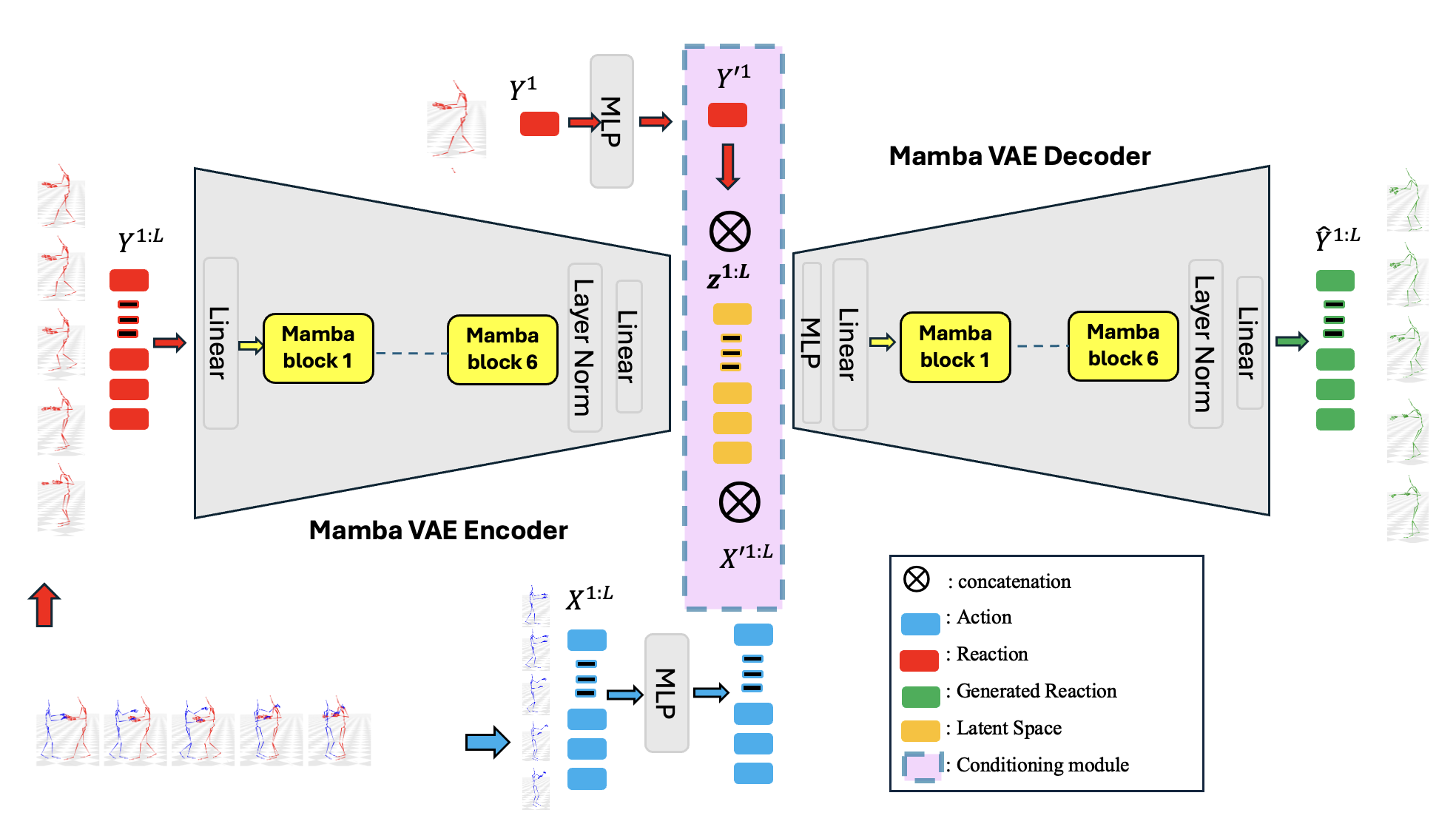}
\vspace{3mm}
\caption{\textbf{VAE Architecture of ReactionMamba.} The Mamba encoder encodes the motion of the reaction $\mathbf{Y}$ and maps it to a latent representation $\mathbf{z}$. The decoder network takes as input the concatenation of the latent vector $\mathbf{z}$, a projection of the action motion sequence $\mathbf{X}$, and the initial pose $Y_1$, and produces the reconstructed reaction motion sequence $\hat{\mathbf{Y}}$.} 
\label{fig:overview}
\end{figure*}

We aim to generate high-quality \textit{reaction motion sequences} that are temporally and semantically consistent with a given \textit{action motion}. The model operates on 3D skeletal pose data, where each pose is a set of joint coordinates over time. 
The goal is to generate a reaction $\mathbf{Y}{=}\{Y_1, Y_2, \dots, Y_T\}$, a sequence of skeleton poses, from $\mathbf{X}{=}\{X_1, X_2, \dots, X_T\}$, a sequence representing the action motion. In this section, we detail the architecture and operational principles of the Reaction Mamba framework, designed for the efficient generation of long-range human reaction motion conditioned on a source action. Initially, we introduce the foundational Mamba model underpinning our approach. We then describe our specialized architecture, which leverages Mamba blocks to improve motion generation efficiency. This framework comprises an encoder for processing input reaction sequences and a decoder that reconstructs the reaction motion.

\subsection{Preliminaries}

\textbf{Selective Structured State Space Models (SSMs)}, specifically recent advancements such as \textit{S4} \cite{gu2022efficiently} and \textit{Mamba} \cite{gu2024mambalineartimesequencemodeling}, have demonstrated superior performance in modeling long-range dependencies. 
Mamba is a selective structured state space model (SSM) that goes beyond the classical discretized linear time-invariant (LTI) state space formulations by introducing input-dependent parameterization, thus enabling information selection over time.

Given an input sequence $\{x_t\}_{t=1}^T$, Mamba has a latent state $h_t \in \mathbb{R}^N$ that evolves according to
\begin{equation}
h_t = \bar{\mathbf{A}}(x_t)\, h_{t-1} + \bar{\mathbf{B}}(x_t)\, x_t ,
\label{eq:mamba-1}
\end{equation}
where $\bar{\mathbf{A}}(x_t)$ and $\bar{\mathbf{B}}(x_t)$ denote discretized and input-conditioned state transition and input projection matrices, respectively. This selectivity allows the model to dynamically select relevant information over time.

The output is computed as
\begin{equation}
y_t = \mathbf{C}(x_t)\, h_t + \mathbf{D}\, x_t ,
\label{eq:mamba-2}
\end{equation}
where $\mathbf{C}(x_t)$ is also an input-dependent output projection and $\mathbf{D}$ represents a learned skip connection that is used for stabilization  and optimization purposes.

The discretization of the continuous-time dynamics is done with the help of an input-dependent time-step parameters $\Delta t(x_t)$, which allow the model to keep or forget information depending on the input. Despite this added complexity, Mamba maintains linear-time complexity $\mathcal{O}(T)$ by using a parallelizable scan operation. This allows the model to be more adapted to the input data and on the same time avoid the quadratic cost of self-attention which allow it to remain effective at modeling long-range dependencies.
As illustrated in \ref{fig:overview}, we integrate Mamba-based modules into both the encoder and decoder to effectively capture long-range temporal relationships within the sequential motion data.

As illustrated in Fig.~\ref{fig:overview}, we integrate Mamba modules into both the encoder and decoder to effectively capture long-range temporal relationships within the sequential motion data.
\subsection{Reaction Mamba}

The Reaction Mamba VAE $\mathcal{V}$ utilizes a Mamba-based architecture \cite{mamba,mamba2} consisting of an encoder $\mathcal{E}$ and a decoder $\mathcal{D}$. We define the mapping and structure of $\mathcal{V} = \{\mathcal{E}, \mathcal{D}\}$ as follows.

The \textbf{encoder network $\mathcal{E}$} processes the input reaction motion sequence $\mathbf{Y}$ by first employing a linear layer to project each pose into a $d_{\text{model}}$-dimensional hidden space. As illustrated in Fig.~\ref{fig:overview}, this projected sequence is processed by a series of Mamba-based modules, following the architecture proposed in \cite{mambaGit}. Each block incorporates a Mamba Selective State Space Model (SSM) and a gated multilayer perceptron (MLP) with $d_{\text{intermediate}}$ hidden units. 

Within each block, residual connections and Root Mean Square (RMS) normalization are applied to stabilize the hidden state transitions. The initial block operates on the projected reaction sequence, while subsequent blocks ingest the hidden states and residuals from their predecessors. After traversing all interconnected blocks, a final normalization step and a linear projection yield the encoder features of dimension $T \times d_{\text{model}}$. This spatio-temporal encoder maps the reaction sequence $\mathbf{Y}$ into a sequence of latent variables $\mathbf{z} = \{\mathbf{z}_1, \dots, \mathbf{z}_T\}$, where $\mathbf{z}_i \in \mathbb{R}^{d_z}$. This is formulated through a variational posterior that produces the mean $\mu$ and variance $\sigma$ used for the reparameterization trick during training.



The \textbf{conditioning module} maps the action sequence $\mathbf{X}$ and the initial reaction pose $\mathbf{Y}_1$ to contextual embeddings $\mathbf{X}' \in \mathbb{R}^{T \times d_a}$ and $\mathbf{Y}_1' \in \mathbb{R}^{T \times d_p}$. These embeddings are subsequently concatenated with the latent vector $\mathbf{z}$ to form the joint representation:

\begin{equation}
\label{equ:concat}
     \mathbf{C} = Concat(\mathbf{z}, \mathbf{X}', \mathbf{Y}_1'),
\end{equation}

where $Concat$ denotes the concatenation operation, resulting in a unified input for the subsequent Mamba-based decoding layers.


The \textbf{decoder network $\mathcal{D}$} reconstructs the reaction motion sequence $\hat{\mathbf{Y}}$ from  $\mathbf{C}$ (\ref{equ:concat}). This concatenated representation is subsequently projected into the hidden dimension $d_{\text{model}}$ through a linear layer.

The decoder architecture mirrors the encoder, utilizing a series of Mamba-based modules. Each block comprises a Mamba layer for temporal modeling followed by a gated MLP, with residual connections and RMS normalization applied at each stage to ensure architectural consistency and training stability. Following the final block, a normalization layer and a linear projection are used to map the hidden states back to the motion space. Formally, the decoder generates the reconstructed reaction $\hat{\mathbf{Y}} \in \mathbb{R}^{T \times d}$ through the following process:
\begin{equation}
\hat{\mathbf{Y}} = \mathcal{D}\left( \text{MLP}(\mathbf{C}) \right),
\end{equation}
More details about the model are given in Section II of Supplementary material.

\color{black}

\subsection{Loss Functions} We use the reconstruction loss, the KL divergence loss, and reaction loss to optimize our model. The \emph{reconstruction loss} (MSE) compares the generated reaction \(\mathbf{\hat{X}} \in \mathbb{R}^{T \times d}\)
 and the ground-truth reaction \(\mathbf{X}\in\mathbb{R}^{T\times d}\). The loss of reconstruction, $\mathcal{L}_{\text{recon}}$, ensures that the generated reaction is perceptually and structurally similar to the generated reaction. The \emph{KL divergence} $\mathcal{L}_{\text{KL}}$ enforces latent variable distributions to follow a prior distribution. The KL divergence loss regularizes the latent space by encouraging it to conform to a prior distribution, ensuring smoothness and continuity in the learned latent representations. Finally, similar to~\cite{ghosh2024remos}, we define \emph{reaction loss} $\mathcal{L}_{\mathrm{react}}$ which captures how closely each predicted reaction remains consistent with the actor’s motion \(\mathbf{Y}\in \mathbb{R}^{T\times d}\). The reaction loss encourages the physical consistency with the action motion and penalizes deviations from the expected motion difference in joint space. It is defined by:
\begin{equation}
\begin{aligned}
\mathcal{L}_{\mathrm{react}} =
\frac{1}{T} \sum_{t=1}^{T}
\exp\!\left( -\|Y_t - X_t\|_2 \right)\times\text{distance} \\
\text{distance}=\left( \|Y_t - \hat{X}_t\|_2 - \|Y_t - X_t\|_2 \right)^2
\end{aligned}
\end{equation}

The total loss function is expressed as:
\begin{equation}
\mathcal{L} = w_{\mathrm{recon}}\mathcal{L}_{\text{recon}} + w_{\mathrm{KL}}\mathcal{L}_{\text{KL}} +  w_{\mathrm{react}}\mathcal{L}_{\text{react}}
\end{equation}

\section{Experiments}

\begin{table*}[t]
\centering
\caption{Comparison of Results Across Different Datasets. \textbf{Bold} denotes the best results, while \underline{underline} indicates the second-best results.}
\vspace{0.5em}

\resizebox{\textwidth}{!}{%
{\scriptsize
\begin{tabular}{l|cccc|cccc}
\hline
& \multicolumn{4}{c|}{\textbf{Lindy Hop (20 Frames $\sim$ 1s)}} 
& \multicolumn{4}{c}{\textbf{NTU120-AS (60 Frames $\sim$ 3s)}} \\
\hline
\textbf{Model} 
& \textbf{MPJPE} & \textbf{MPJVE} & \textbf{$FID$} & \textbf{$div_{gen}/div_{gt}$}
& \textbf{MPJPE} & \textbf{MPJVE} & \textbf{$FID$} & \textbf{$div_{gen}/div_{gt}$} \\
\hline

ReMoS~\cite{ghosh2024remos} 
& 0.2194 & \underline{0.0033} & \underline{2.3097} & \textbf{1.0021}
& 0.4577 & 0.0030 & 26.859 & 0.9654 \\

InterFormer~\cite{ChopinInterformer2023} 
& 0.1723 & 0.0046 & 2.7885 & 0.9936
& \underline{0.1865} & \textbf{0.0020} & 5.9255 & 1.0218 \\

Ready-to-React~\cite{ready_to_react_repo}  
& \textbf{0.1483} & \textbf{0.0031} & 25.711 & 1.0032
& 0.5006 & 0.0037 & 12.292 & \textbf{0.9872} \\

Seq2SeqMamba
& 0.2463 & 0.0056 & 3.0005 & \underline{1.0011}
& 0.1942 & 0.0027 & \underline{0.8256} & 0.9784 \\

\rowcolor{gray!15} ReactionMamba (Ours) 
& \underline{0.1635} & 0.0037 & \textbf{1.9852} & 1.0025
& \textbf{0.0922} & \textbf{0.0020} & \textbf{0.0321} & \underline{0.9816} \\
\hline
\end{tabular}
}}
\label{tab:fused_results}
\end{table*}

\subsection{Datasets}
We validated the proposed method in a broad set of experiments on 3 publicly available interaction datasets with 3D skeletons.\\
\noindent
\textbf{ReMoCap}~\cite{ghosh2024remos} dataset consists of two types of two-person interactions: Lindy Hop dancing and Ninjutsu martial arts. 
\textbf{The Lindy Hop}~\cite{ghosh2024remos} subset comprises 8 dance motion sequences, each approximately 7.5 minutes long and recorded at 50 frames per second. We tested our method and the baselines on this subset of ReMoCap following the preprocessing and train/test split provided by the authors.
\textbf{NTU120-AS}~\cite{xu2024regennet} is an annotated and extended version of the NTU RGB+D 120 dataset \cite{Liu2020NTURGBD120}. It contains 26 action categories, such as “kick”, “push”, “hug”, and “finger guessing”.
 We used the same data processing as in \cite{xu2024regennet}.
\textbf{InterX}~\cite{xu2024inter} is a large-scale dataset comprising 11,388 interaction sequences across 40 categories, such as “handshake”, “hug”, “pull”, and “point”. Each sequence is annotated with actor–reactor roles. For comparisons on \textbf{ReMoS}~\cite{ghosh2024remos}, \textbf{Interformer}~\cite{ChopinInterformer2023}, Seq2SeqMamba, and ReactionMamba, we use joint data in a 3D coordinate system provided by the official database, apply downsampling, and construct a custom train/test split. For comparisons involving \textbf{Think-Then-React}~\cite{tan2025think}, we follow the official data split and preprocessing protocol released by the authors~\cite{tan2025think}. Additional details on data sampling and preprocessing are provided in Section 8 of the Supplementary Material. 

\subsection{Evaluation Metrics}

To evaluate the fidelity and diversity of the generated reaction sequences $\hat{\mathbf{Y}}$ relative to the ground truth $\mathbf{Y}$, we employ four quantitative metrics on the normalized data. First, the \textbf{Mean Per Joint Position Error (MPJPE)} measures spatial accuracy by computing the average Euclidean distance between corresponding joint coordinates. Second, to ensure temporal consistency, the \textbf{Mean Per Joint Velocity Error (MPJVE)} assesses the average difference in joint velocities between predicted and ground truth sequences. Third, the overall quality and realism of the motion are quantified via the \textbf{Fréchet Inception Distance (FID)}~\cite{GANFIDNIPS2017}, which compares the distribution of high-level features extracted from real and generated sequences. Finally, \textbf{Diversity}~\cite{dancingmusic,zhang2018perceptual} is calculated to measure the variability within the generated set, ensuring the model captures a broad range of reactive behaviors.

\begin{table}[!htbp]
\scriptsize
\centering
\caption{Comparison of long sequence generation on InterX (200 frames $\sim$10s) under the \textbf{57-joint setting}.
\textbf{Bold} denotes the best and \underline{underline} the second-best results.}
\vspace{0.5em}
\begin{tabular}{l|cccc}
\hline
\textbf{Model} 
& \textbf{MPJPE} 
& \textbf{MPJVE} 
& \textbf{FID} 
& \textbf{$div_{gen}/div_{gt}$} \\
\hline
ReMoS~\cite{ghosh2024remos} 
& 0.3318 & 0.0033 & 51.272 & 0.7891 \\

InterFormer~\cite{ChopinInterformer2023} 
& 0.2394 & 0.0025 & 6.1582 & 0.9329 \\

Seq2SeqMamba 
& \underline{0.0483} & \textbf{0.0006} & \underline{0.4055} & \underline{0.9966} \\

\rowcolor{gray!15}
ReactionMamba (Ours, 57J) 
& \textbf{0.0471} & \underline{0.0007} & \textbf{0.0477} & \textbf{1.0026} \\
\hline
\end{tabular}
\label{tab:interx_57j}
\end{table}

\subsection{Baselines}
We evaluate our approach against four state-of-the-art methods: \textbf{ReMoS}~\cite{ghosh2024remos}, \textbf{Interformer}~\cite{ChopinInterformer2023}, \textbf{Ready-to-React}~\cite{cen2025_ready_to_react} and \textbf{Think-Then-React}~\cite{tan2025think}.
In addition, we include a simple \textbf{Seq2SeqMamba}~\cite{mamba} baseline to isolate the contribution of the Mamba state-space backbone.

The \textbf{ReMoS}~\cite{ghosh2024remos} framework employs a decoupled architecture consisting of independent body and hand diffusion modules. We retrained both modules on each of the three evaluation datasets, strictly following the implementation guidelines provided by the authors~\cite{anindita127ReMoS}.

\textbf{Interformer}~\cite{ChopinInterformer2023}, a Transformer-based architecture for reaction synthesis, was retrained on all datasets adhering to the experimental protocol established in~\cite{ChopinInterformer2023, ghosh2024remos}.
\textbf{Ready-to-React}~\cite{cen2025_ready_to_react} uses an online reaction policy that conditions motion generation on the historical trajectories of both interacting agents to enable context-aware responses. In our experiments, we use its reaction generation mode to predict an agent's future movement based on its past states and the lead agent's motion. However, due to unresolved preprocessing inconsistencies encountered when following the original implementation~\cite{ready_to_react_repo}, we exclude Ready-to-React results for the InterX dataset to maintain the integrity and fairness of the comparative analysis.

\textbf{Think-Then-React}~\cite{tan2025think} is a multi-steps framework that combines a VQ-VAE (Vector-Quantized Variational Autoencoder)~\cite{VQAE2017} motion tokenizer and a language-model-based reasoning function for action-to-reaction generation. We use the official checkpoints and evaluation protocol on InterX without retraining to ensure a fair comparison.

The \textbf{Seq2SeqMamba}~\cite{gu2024mambalineartimesequencemodeling} baseline is a deterministic sequence-to-sequence model based on the Mamba architecture. It directly maps the observed action sequence to the corresponding reaction sequence, without latent variables, stochastic sampling, or explicit conditioning on the initial reaction pose. This baseline is used to assess whether a strong temporal backbone alone is sufficient for reaction motion synthesis.

\subsection{Implementation}

Reaction Mamba was trained using Adam optimizer. The base learning rate is set to $10^{-4}$, it decays following a Cosine Annealing LR scheduler. Supposing $k$ is the number of joints, the input dimension is set to $d =k \times 3$, $d_{model} = 256$, $d_{z} = 128, d_{intermediate} =  512 $. We use 6 sequential layers of the mamba block for both encoder and decoder. Before concatenation, the action and initial frame are projected to $d_{c} = 64$ each of them.  We set the loss weights, $w_{\mathrm{recon}} =  w_{\mathrm{recon}} = 1$ and $w_{\mathrm{KL}} = 0.5$. Reaction Mamba was trained on NVIDIA Quadro RTX 6000 GPU (24 GB VRAM), for 300k iterations for Lindy Hop and 90K iterations for NTU120-AS and InterX more training details are present in Section~10 of the Supplementary Material.


\section{Results}
In this section, we present quantitative and qualitative results pertaining to reaction generation as well as a study of inference time. In each experiment, all models were trained on the dataset they are tested on. Each train set and test set was normalized following the definition in Section~8 of the Supplementary Material.\\
\textbf{Quantitative Results.}
Table~\ref{tab:fused_results} reports results on LindyHop and NTU120-AS using each dataset’s default sequence length. ReactionMamba is competitive with the state of the art on LindyHop across all metrics. Ready-to-React achieves the lowest MPJPE and MPJVE on these short sequences, reflecting strong short-horizon accuracy, but its substantially higher FID indicates reduced realism and temporal consistency. On NTU120-AS, ReactionMamba outperforms all baselines. This can be attributed to the absence of explicit contact annotations in this dataset, which are used by ReMoS but are not required by ReactionMamba, as well as to our model’s ability to handle sequences where the reactor remains near its initial position (e.g., talking while standing up face to face). This behavior is encouraged by injecting the initial pose information at every timestep in the conditioning module.

Table~\ref{tab:interx_57j} presents long-term generation results on InterX with 200-frame sequences under the 57-joint setting. ReactionMamba achieves the best performance across all metrics, producing accurate, smooth, and diverse long-horizon reactions. In contrast, ReMoS exhibits a strong degradation in performance, largely due to the high memory cost of its spatio-temporal cross-attention, which required a reduced batch size even on an NVIDIA H100 SXM5 GPU (80~GB HBM3), limiting its effectiveness.

Table~\ref{tab:interx_22j} further evaluates long-sequence generation under the 22-joint ego-centric setting on InterX, using variable-length sequences ranging from 32 to 256 frames. Under identical joint representations and evaluation protocols, ReactionMamba significantly outperforms Think-Then-React across all metrics, reducing MPJPE by more than a factor of two and achieving substantially lower FID. These results highlight ReactionMamba’s robustness across skeleton configurations and sequence lengths.

Finally, the Seq2SeqMamba baseline consistently underperforms ReactionMamba across all datasets, indicating that a strong temporal backbone alone is insufficient for high-quality and long-term reaction synthesis.

\begin{table}[!htbp]
\small
\centering
\caption{Comparison under the 22-joint setting on InterX (32 to 256 frames $\sim 1.5$ to $13s$).
Both methods are evaluated using the same ego centric joint representation}
\vspace{0.5em}
\resizebox{\linewidth}{!}{

\begin{tabular}{l|cccc}
\hline
\textbf{Model} 
& \textbf{MPJPE} 
& \textbf{MPJVE} 
& \textbf{FID} 
& \textbf{$div_{gen}/div_{gt}$} \\
\hline
Think-Then-React~\cite{tan2025think} (22J) 
& 0.4732 & 0.0021 & 13.675 & 1.2237 \\

\rowcolor{gray!15}
ReactionMamba (Ours, 22J) 
& \textbf{0.1888} & \textbf{0.0019} & \textbf{2.079} & \textbf{0.9328} \\
\hline
\end{tabular}}
\label{tab:interx_22j}
\end{table}

\noindent\textbf{Inference Time.}
Tables~\ref{tab:inference_time} and~\ref{tab:inference_time_22j} report inference speed on an NVIDIA Quadro RTX 6000. On NTU120-AS (Table~\ref{tab:inference_time}), ReactionMamba is much faster than all baselines, being several hundred times faster than Ready-to-React and InterFormer, and more than 14K faster than ReMoS. This highlights the efficiency of its Mamba-based design, which avoids costly spatio-temporal attention.

Under the 22-joint ego-centric setting on InterX (Table~\ref{tab:inference_time_22j}), ReactionMamba again substantially outperforms Think-Then-React, achieving real-time generation even for variable-length sequences up to 256 frames. Overall, these results demonstrate that ReactionMamba scales efficiently to large test sets while enabling fast and practical long-horizon reaction generation.

\begin{table}[!ht]
\small
\centering
\caption{Inference Speed on NVIDIA Quadro RTX 6000 for 3840 NTU-AS120 Sequences of 60 Frames  each}
\resizebox{\linewidth}{!}{
\begin{tabular}{l|ccc}
\hline
\textbf{Model} & \textbf{Total Time (min)} & \textbf{Time per Sequence (s)} & \textbf{FPS} \\
\hline
ReMoS~\cite{ghosh2024remos}        &1691    &  26  & 2.3 \\
Interformer~\cite{ChopinInterformer2023}        & 122 &1.9  & 32\\

Ready-to-React~\cite{ready_to_react_repo}        & 105 &  1.6& 38\\

\rowcolor{gray!15} ReactionMamba (Ours)  & \textbf{0.117} & \textbf{0.0018} & $\mathbf{33000}$\\
Seq2SeqMamba & 0.07 & 0.000018& 330 000 \\
\hline
\end{tabular}}
\label{tab:inference_time}
\end{table}

\begin{table}[!ht]
\small
\centering
\caption{Inference speed on NVIDIA Quadro RTX 6000 under the 22-joint setting on InterX.
Results are reported on 1632 test sequences with motion length between 32 and 256 frames.}
\vspace{0.5em}
\resizebox{\linewidth}{!}{
\begin{tabular}{l|ccc}
\hline
\textbf{Model} 
& \textbf{Total Time (min)} 
& \textbf{Time per Sequence (s)} 
& \textbf{FPS} \\
\hline
Think-Then-React~\cite{tan2025think} & 11,28 & 0,41 &  265 \\

\rowcolor{gray!15}
ReactionMamba (Ours, 22J)  & 0.10 & 0.0037 & 30025 \\
\hline
\end{tabular}}
\label{tab:inference_time_22j}
\end{table}



\begin{figure*}[!htbp]
    \vspace*{-1.5cm}
    \centering
    \includegraphics[width=0.93\textwidth]{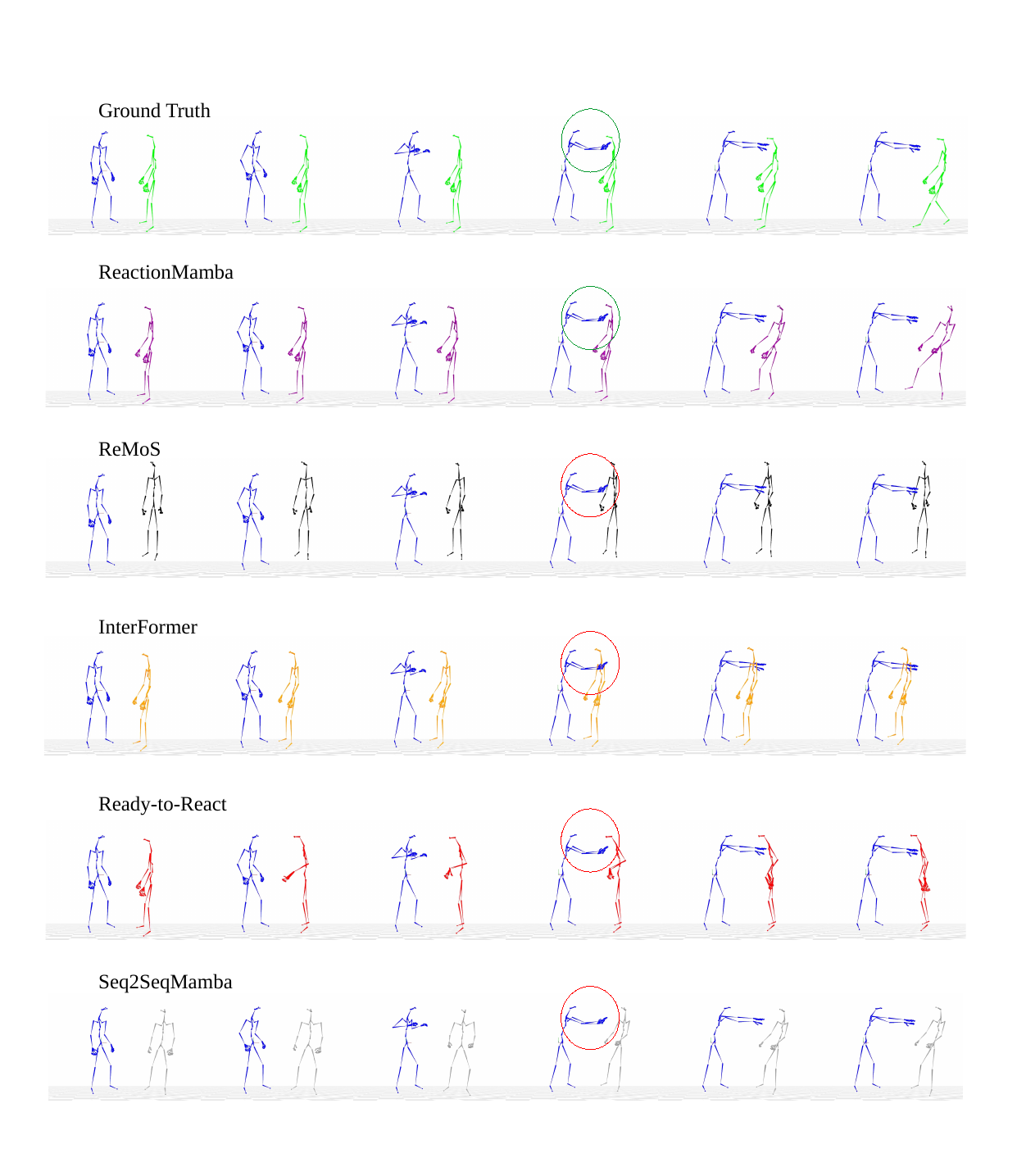}
    \caption{ Visualization of sequence generated on \textbf{NTU120-AS push class}. In \textcolor{blue}{blue} the action motion used as condition. In \textcolor{green}{green} the ground truth reaction and in other colors the reaction generated by the different models. See the Supplementary Material for their corresponding animations.}
    \label{fig:ntuas}
\end{figure*}

\noindent\textbf{Qualitative Results.} We visualize reaction generation on NTU120-AS for the \emph{push} class (60 frames) and refer to the supplementary material for extended long-horizon results on LindyHop (1000 frames). As illustrated in Fig.~\ref{fig:ntuas}, ReactionMamba produces more physically consistent and realistic reactions to being pushed. In comparison, other methods tend to exhibit reduced motion intensity and occasional body interpenetrations, highlighted by the green circles (plausible gestures) and red circles (failure cases).

Overall, the qualitative results are consistent with the quantitative findings. ReactionMamba closely follows the ground-truth reaction while maintaining stable and coherent interactions. InterFormer and Seq2SeqMamba struggle to produce meaningful hand contacts, while Ready-to-React shows noticeable drift after the initial frames, resulting in less coherent and poorly articulated motions. Additional long-term qualitative results on LindyHop are provided in the supplementary material (Fig.~5), further illustrating ReactionMamba’s stability and realism over long-term generation.

\noindent\textbf{User Study.}  The user study compared \textbf{ReactionMamba} with three leading methods: ReMos~\cite{ghosh2024remos}, InterFormer~\cite{ChopinInterformer2023}, and Ready-to-React~\cite{ready_to_react_repo}. For each dataset, results generated by ReactionMamba, Interfomer, ReMoS, and Ready-to-React are included. For the same interaction class, the results are randomly presented to users, who are asked to answer two questions: (1) Motion Quality, which evaluates the overall perceptual quality of the generated reaction motion in terms of smoothness, naturalness, and physical plausibility (1 = very low quality, 7 = very high quality); and (2) Coherence with the Action, which measures how coherent the generated reaction motion is with respect to the action motion (1 = not coherent at all, 7 = highly coherent). A total of 15 users participate in the evaluation, and the final results are reported as the average scores across all users.


Scores are reported as mean opinion scores (MOS), averaged over 27 interaction prompts. Overall, the results in Table~\ref{tab:user_study_mos} demonstrate that our method produces higher-quality reaction motion sequences than the competing approaches.

\begin{table}[!htbp]
\centering
\caption{User study results (Mean Opinion Score, MOS) on reaction motion generation.
Scores are averaged over 27 interaction prompts. Higher is better.}
\label{tab:user_study_mos}
\setlength{\tabcolsep}{6pt}
\renewcommand{\arraystretch}{1.2}
\begin{tabular}{l|cc|cc|cc}
\hline
\multirow{2}{*}{\textbf{Method}}
& \multicolumn{2}{c|}{\textbf{Lindy Hop}}
& \multicolumn{2}{c|}{\textbf{NTU120-AS}}
& \multicolumn{2}{c}{\textbf{InterX}} \\
\cline{2-7}
& \textbf{Q1} & \textbf{Q2}
& \textbf{Q1} & \textbf{Q2}
& \textbf{Q1} & \textbf{Q2} \\
\hline

ReMoS~\cite{ghosh2024remos}               & 3.6 & 3.8 & 4.0 & 4.1 & 3.9 & 4.2 \\
InterFormer~\cite{ChopinInterformer2023}  & 4.0 & 4.2 & 4.4 & 4.5 & 4.4 & 4.8 \\
Ready-to-React~\cite{ready_to_react_repo} & 4.5 & 5.3 & 4.8 & 5.2 & - & - \\
\rowcolor{gray!15}  ReactionMamba (Ours)                      & \textbf{5.8} & \textbf{6.2} & \textbf{5.7} & \textbf{6.3} & \textbf{6.1} & \textbf{6.4} \\
\hline

\end{tabular}
\end{table}

\noindent\textbf{Ablation.}
We perform an ablation study to evaluate the influence of different components within our proposed ReactionMamba framework. We report ablation results in Table ~\ref{tab:ablation_lh} for LindyHop (20 frames) 
Five VAE variants are studied. 
\textbf{S1 (VAE + Mamba + Concat Action + Concat Init-Pose)} corresponds to our final model, where both the action stream and the initial pose are incorporated into the decoder via direct concatenation.
\textbf{S2 (S1 $\rightarrow$ Replace Mamba with Attention)} keeps the same architecture as S1 but replaces all Mamba state space layers in the encoder and decoder with attention layers, which leads to 
degradation, especially in FID, confirming the effectiveness of Mamba. 
\textbf{S3 (S1 $\rightarrow$ Remove Init-Pose Concatenation)} removes the initial pose stream entirely in order to underline its necessity for generating consistent reaction motions.
\textbf{S4 (S1 $\rightarrow$ Replace Init-Pose Concatenation with Gating)} introduces a 
time-decayed gating mechanism, where the initial pose is injected into the latent sequence such that its influence gradually fades over time. While gating provides competitive performance, it consistently underperforms concatenation in reconstruction accuracy and stability, showing that directly concatenating the initial pose at every frame 
remains a more reliable conditioning signal.
Finally, \textbf{S5 (S1 $\rightarrow$ Replace Action Concatenation with Cross-Attention)} replaces direct concatenation of the action stream with a cross-attention mechanism that enriches the latent sequence with action information.
While cross-attention produces slightly improved distributional realism compared to removing the initial pose or replacing Mamba with attention (S2 and S3), it comes at the cost of higher reconstruction errors, indicating that action concatenation remains the more balanced and robust strategy overall. \textbf{S6 (S1 → Replace Action Concatenation with Addition)} injects the projected action features into the latent sequence via element-wise addition.

To conclude, for short sequences, concatenation clearly dominates (S1), achieving the best performance across all evaluation metrics, including reconstruction accuracy, distributional realism, and diversity. While gating (S4) or cross-attention (S5) or addition (S6) can offer competitive results on individual metrics, neither variant surpasses the overall balance provided by S1. We therefore adopt S1 (VAE + Mamba + Concat(Action) + Concat(Init-Pose)) as our final architecture, and report S4 and S5 as targeted alternatives when one metric (MPJPE or FID) is prioritized over the global trade-off.

\begin{table}[!htbp]
\centering
\begin{tabular}{l|cccc}
\toprule
Scenario & MPJPE ↓ & MPJVE ↓ & FID ↓ & \textbf{\(div_{gen}/div_{gt}\)}  \\
\midrule
S1  & \textbf{0.1634} & \textbf{0.00374} & \textbf{1.9852} & \textbf{1.0025}     \\
S2  & 0.1833 & 0.00407 & 2.2392 & 1.0037     \\
S3  & 0.2299 & 0.00543 & 2.3857 & 1.0157   \\
S4  & 0.1773 & 0.00410 & 2.0318 & 0.9836     \\
S5  & 0.2067 & 0.00427 & 2.0915 & 0.9925   \\
S6  & 0.1708 & 0.00388 & 2.1890 & 0.9943 \\
\bottomrule
\end{tabular}
\caption{Ablation study on ReactionMamba. Results for LindyHop sequences of 20 frames each}
\label{tab:ablation_lh}
\end{table}

\section{Limitations and future work}
\label{sec:discussion}

While ReactionMamba remains competitive with state-of-the-art methods and is considerably faster than previous approaches, it still exhibits certain limitations.
Injecting initial pose information at each frame into the conditioning module generally improves generation quality; however, the conditioning can sometimes be overly strong, leading to incorrect reactions (e.g., when the initial pose closely resembles that of a specific reaction) or even a lack of motion. In addition, like many existing models, ReactionMamba occasionally struggles to produce realistic leg movements, causing the character to “glide” rather than move naturally—particularly on the NTU120-AS dataset, where most actions and reactions are performed in place.
In the future, to address these issues, we plan to adaptively weight the initial pose conditioning based on the type of action performed and its temporal progression. For instance, actions such as “Follow the other person” in NTU120-AS require the model to move significantly away from the initial pose. Thus, for such actions, the initial pose could be assigned greater importance at the beginning of the motion and gradually reduced over time.
Furthermore, foot movement realism may be improved by incorporating a foot contact loss as proposed in~\cite{ghosh2024remos, tevet2023human}.


\section{Conclusion}
We present ReactionMamba, a new model for reaction generation. ReactionMamba adopts a VAE structure with an encoder and decoder based on the Mamba architecture. Extensive experiments on four datasets show that ReactionMamba is competitive with state-of-the-art methods both quantitatively and qualitatively for short-term and long-term reaction generation, while being up to 1000 times faster. This substantial increase in generation speed enables real-time use of ReactionMamba and opens a wider range of applications than existing methods.

\section*{ETHICAL IMPACT STATEMENT}
Our assessment did not reveal any foreseeable risks or adverse impacts associated with ReactionMamba. Therefore, no specific risk‑mitigation measures were deemed necessary.
\section*{Acknowledgements}
This project is supported by a French government grant under the
France 2030 program – ANR-22-EXEN-0004 (PEPR eNSEMBLE / PC3).

%
%


\section*{Supplementary Materials}
In this supplementary document, we provide details about the data processing pipeline, provide the implementation details of our experiments, and also a snapshot of the user study interface.

\section{Data Preprocessing}

We evaluate our method on three datasets: \textbf{InterX}, \textbf{NTU120-AS}, and \textbf{Lindy Hop}. 
To ensure a fair comparison across datasets and with prior work ~\cite{ghosh2024remos}, we apply a unified pre processing pipeline to all three datasets.

\subsection{Skeleton Representation}
\par All motion sequences are represented using \textbf{3D joint positions only}, expressed in Cartesian coordinates $(x, y, z)$. 

All datasets three include both body and hand joints with:
\begin{itemize}
    \item \textbf{Lindy Hop}: 47 joints (body + hands).
    \item \textbf{NTU120-AS}: 55 joints (body + hands).
    \item \textbf{InterX}: 51 joints (body + hands).
\end{itemize}
\par For comparisons with Think-Then-React (TTR) on InterX, we use a reduced \textbf{22-joint body-only representation}, consistent with the official TTR preprocessing~\cite{tan2025think}. 
This representation excludes hand joints and focuses on the torso and limbs, enabling direct comparison with the released Think-Then-React checkpoints.
All other experiments use the full body+hand representations described above.

\subsection{Sequence Length and framerate}
All datasets are sampled at a rate of \textbf{20 FPS}. 
During training, each model is trained on fixed-length motion sequences:

\begin{itemize}
    \item \textbf{InterX}: 200 frames per sequence.
    \item \textbf{NTU120-AS}: 60 frames per sequence.
    \item \textbf{Lindy Hop}: 20 frames per sequence.
\end{itemize}

Although training is performed on fixed-length sequences, longer reaction motions can be generated at inference time using an auto-regressive strategy. Specifically, the model generates a fixed number of frames, and the last frame of one subsequence is used as the initial pose for the next subsequence.

\subsection{Sequence Sampling Method}
For the sampling we following prior work:

\begin{itemize}
    \item \textbf{Lindy Hop}: Since the dataset does not provide action class annotations, we adopt a sliding-window sampling strategy, following the preprocessing used in the ReMoS codebase~\cite{ghosh2024remos}.
    
    \item \textbf{NTU120-AS}: We follow the same preprocessing protocol as ReGenNet~\cite{xu2024regennet}, extracting fixed-length segments from each labeled sequence.
    
    \item \textbf{InterX}: Sequences are processed one by one following InterX~\cite{xu2024inter} and we decided to select 200 frames for each to capture long interactions .
\end{itemize}

\subsection{Data Normalization}
For a fair comparison, all sequences from Lindy Hop, NTU120-AS, and InterX are normalized using the same technique introduced in the ReMoS paper~\cite{ghosh2024remos}. 
This normalization was used not only for the ReMoS model itself but also for all baseline methods evaluated in that work execept Think-Then-React~\cite{tan2025think}. 
Since ReMoS is one of our primary baselines, we took inspiration from it applied this normalization uniformly across all three datasets for each model.

\paragraph{Definitions}
\begin{itemize}
  \item $\mathbf{P}^{(i)} \in \mathbb{R}^{T \times J \times 3}$: global joint positions of character $i$.
  \item $\mathbf{r}^{(1)} = \mathbf{P}^{(1)}[:, 0] \in \mathbb{R}^{T \times 3}$: root trajectory of the actor (joint index 0).
  \item $s \in \mathbb{R}$: scale factor.
\end{itemize}

\paragraph{Body Normalization}
Body joints are normalized by subtracting the actor’s root trajectory and applying scale normalization:
\begin{equation}
\hat{\mathbf{P}}^{(i)}_{\text{body}} =
\frac{\mathbf{P}^{(i)}_{\text{body}} - \mathbf{r}^{(1)}}{s}
\end{equation}

\paragraph{Hand Normalization}
Hand joints are normalized relative to the wrist joint:
\begin{equation}
\mathbf{w}^{(i)}_{\text{rh}} = \mathbf{P}^{(i)}_{\text{rh}}[:, 0], \quad
\mathbf{w}^{(i)}_{\text{lh}} = \mathbf{P}^{(i)}_{\text{lh}}[:, 0]
\end{equation}
\begin{equation}
\hat{\mathbf{P}}^{(i)}_{\text{rh}} =
\frac{\mathbf{P}^{(i)}_{\text{rh}} - \mathbf{w}^{(i)}_{\text{rh}}}{s}, \quad
\hat{\mathbf{P}}^{(i)}_{\text{lh}} =
\frac{\mathbf{P}^{(i)}_{\text{lh}} - \mathbf{w}^{(i)}_{\text{lh}}}{s}
\end{equation}
\paragraph{Ego-Centric Representation (InterX, Think-Then-React vs ReactionMamba Comparison)}
For comparisons involving Think-Then-React ~\cite{tan2025think} on the InterX dataset, we adopt the \textbf{ego-centric representation} introduced in~\cite{tan2025think}. 
In this formulation, the reactor defines the reference coordinate system: the reactor’s root joint at the first frame is translated to the origin on the ground plane, and the reactor is rotated to face the positive Z-axis. 
Both the actor and reactor motions are then expressed in this shared ego-centric coordinate system.

This ego-centric normalization differs from the actor-centered normalization used in ReMoS~\cite{ghosh2024remos} and is applied exclusively for experiments involving Think--Then--React to ensure a faithful comparison with the original method.

This representation differs from the actor-centered normalization used in ReMoS~\cite{ghosh2024remos} and is only applied for experiments involving Think--Then--React to ensure a faithful and fair comparison.

\section{Model Details}

ReactionMamba follows a conditional variational autoencoder (VAE) formulation. It is made of three main components: a Mamba-based encoder, a conditioning module, and a Mamba-based decoder. All architectural hyperparameters were selected empirically after evaluating multiple configurations and choosing the ones that provided the best performance

\subsection{Encoder and Latent Space}
\begin{figure*}[!ht]
    \centering
    \includegraphics[width=0.9\linewidth]{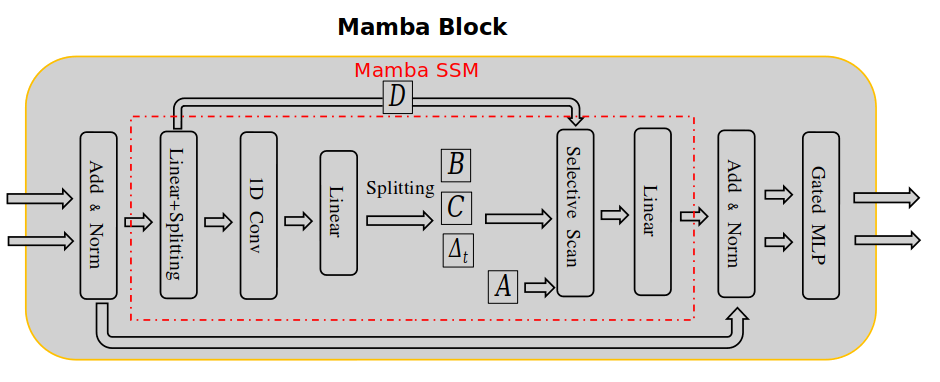}
    \caption{Focus on a Mamba block. The red component represents the state update governed by Eq.~(1) and Eq.~(2) in the main paper.}
    \label{fig:zoom}
    \vspace{-0.4cm}
\end{figure*}

The encoder backbone consists of 6 sequential Mamba blocks, as shown in Fig. 2 in the main paper with a hidden dimension of 256. Each block uses an intermediate feed-forward network with a dimension of 512 and is represented in Fig.~\ref{fig:zoom}.

During training, the encoder processes the ground-truth reaction motion sequence and produces per-time-step parameters (mean and variance) of a Gaussian latent distribution. The latent dimension is set to 128. Latent variables are sampled using the standard reparameterization trick. 

\subsection{Conditioning Design}
Reaction generation is conditioned on two sources of information: (i) the full actor motion sequence and (ii) the initial pose of the reactor.

Both conditioning signals are encoded using lightweight per-frame MLPs that project the flattened joint positions into a 64-dimensional embedding space. Each MLP consists of a linear layer followed by a GELU activation and Layer Normalization. The initial reactor pose is extracted as the first reaction frame and repeated across time to form a conditioning sequence.

The latent representation, actor motion embedding, and initial pose embedding are concatenated and jointly provided as input to the decoder.

\subsection{Decoder}
The decoder mirrors the encoder architecture and consists of 6 stacked Mamba layers with a hidden dimension of 256. The concatenated latent and conditioning representation, which has a total dimension of $128 + 64 + 64 = 256$, is first projected to the decoder input dimension and then decoded into per-frame 3D joint positions.

\subsection{Inference}
At inference time, the encoder is not used. Latent variables are sampled from a standard normal distribution, and the decoder generates the reaction motion conditioned on the actor motion and the initial reactor pose.
Although training is performed on fixed-length sequences, longer reaction motions are generated autoregressively by chaining subsequences, using the last generated frame of one subsequence as the initial pose for the next.

\section{Training parameters}

Table~\ref{tab:final_training_config} summarizes the training configurations for ReMoS, InterFormer, and ReactionMamba. For ReMoS, the target number of iterations and batch size are derived from the training schedules implied by the original paper~\cite{ghosh2024remos} and the released checkpoints. For the Lindy Hop dataset, models are trained for a larger number of updates due to its greater dataset size. InterFormer follows a similar dataset-dependent allocation strategy, assigning more iterations to Lindy Hop and fewer to smaller datasets such as NTU120-AS and InterX. ReactionMamba is a VAE-based model and is therefore trained with iteration budgets aligned with InterFormer to ensure a fair and consistent comparison.
To ensure comparability, Seq2SeqMamba was trained using the same configuration settings as ReactionMamba.

\medskip
\noindent
For Ready-to-React, we strictly follow the original training protocol provided in the official implementation. Specifically, we reuse the released codebase without modifying the number of epochs, optimization hyperparameters, or learning rate schedules. The only changes involve replacing the dataset loader and adjusting the number of frames to accommodate our datasets and normalization scheme.

\medskip
\noindent
For comparisons with Think-Then-React on InterX, we follow the official inference protocol released by~\cite{tan2025think} and use their provided pretrained checkpoints without retraining. To ensure a fair comparison, we train ReactionMamba under the same ego-centric setting, using identical training files, representations, and test splits for both methods.

\begin{table*}[t]
\centering
\caption{Training configuration summary for ReMoS, Interformer and ReaactionMamba. }
\label{tab:final_training_config}
\begin{tabular}{llcccc}
\toprule
\textbf{Model} & \textbf{Dataset} & \textbf{Part} & \textbf{\# Training Sequences} & \textbf{Batch Size} & \textbf{Target Iterations} \\
\midrule
ReMoS & Lindy Hop  & Body & 50,328 & 32 & 900k \\
ReMoS & Lindy Hop  & Hand & 50,328 & 64 & 600k \\
ReMoS & NTU120-AS  & Body & 4,224 & 32 & 300k \\
ReMoS & NTU120-AS  & Hand & 4,224 & 64 & 170k \\
ReMoS & InterX & Body & 8,770 & 8 & 300k \\
ReMoS & InterX & Hand & 8,770 & 4 & 170k \\
\midrule
InterFormer & Lindy Hop  & Body+Hand & 50,328 & 128 & 300k \\
InterFormer & NTU120-AS  & Body+Hand & 4,224 & 128 & 90k \\
InterFormer & InterX & Body+Hand & 8,770 & 128 & 90k \\
\midrule
ReactionMamba (Ours) & Lindy Hop  & Body+Hand & 50,328 & 32 & 300k \\
ReactionMamba (Ours) & NTU120-AS  & Body+Hand & 4,224 & 16 & 90k \\
ReactionMamba (Ours) & InterX & Body+Hand & 8,770 & 16 & 90k \\
\bottomrule
\end{tabular}
\end{table*}


\section{Metric Definitions}
To assess the fidelity and diversity of the generated reaction sequences $\hat{\mathbf{Y}}$ compared to the ground truth $\mathbf{Y}$, we compute several quantitative metrics on the normalized data. 

\paragraph{Mean Per Joint Position Error (MPJPE)}  MPJPE computes the average Euclidean distance between corresponding joints in the predicted and ground truth sequences. It is defined as:

\begin{equation}
\mathrm{MPJPE} = \frac{1}{T J} \sum_{t=1}^{T} \sum_{j=1}^{J} \left\| \hat{Y}_{t,j} - Y_{t,j} \right\|_2,
\end{equation}
where $T$ is the number of frames, and $J$ is the number of joints. This metric evaluates the spatial accuracy of the predicted poses.

\paragraph{Mean Per Joint Velocity Error (MPJVE)} MPJVE computes the average difference in joint velocities between the predicted and ground truth sequences, capturing temporal consistency:

\begin{align}
\mathrm{MPJVE} = \; & \frac{1}{(T - 1) J} \sum_{t=1}^{T-1} \sum_{j=1}^{J} \nonumber \\
& \left\| (\hat{Y}_{t+1,j} - \hat{Y}_{t,j}) - (Y_{t+1,j} - Y_{t,j}) \right\|_2.
\end{align}


\paragraph{Fréchet Inception Distance (FID)}
FID allows to evaluate the quality of generated data by comparing the distribution of features extracted real and generated motion sequences using a motion encoder. After assuming these features follow multivariate Gaussian distributions with means and covariances $(\mu, \Sigma)$ and $(\hat{\mu}, \hat{\Sigma})$ respectively, FID is computed as:
\begin{equation}
\mathrm{FID} = \left\| \mu - \hat{\mu} \right\|_2^2 + \mathrm{Tr}\left( \Sigma + \hat{\Sigma} - 2(\Sigma \hat{\Sigma})^{1/2} \right).
\end{equation}

A lower FID indicates that the generated data distribution closely matches the real data distribution, reflecting higher quality and diversity in the generated motions ~\cite{FrechetDistance,GANFIDNIPS2017}.
For all methods, FID is computed using the same transformer based motion encoder trained on the corresponding dataset and joint representation.

\paragraph{Diversity Metrics}
To specifically assess the diversity of the generated motions, we randomly sample pairs of motion sequences from feature space and compute the average Euclidean distance between them across frames and joints:
\begin{equation}
\mathrm{Diversity}(S) = \frac{1}{|P|} \sum_{(i, j) \in P} \frac{1}{T J} \sum_{t=1}^{T} \sum_{j=1}^{J} \left\| S^i_{t,j} - S^j_{t,j} \right\|_2,
\end{equation}
where $S^i$ and $S^j$ are two sampled motion sequences from set $S$ (real or generated), and $P$ is the set of unique pairs randomly selected. A diversity score close to the ground truth diversity score reflects greater variability within the motion set and preservation of the original diversity.

\section{More visual results}

\begin{figure*}[!ht]
    \centering
    \includegraphics[width=1.0\textwidth]{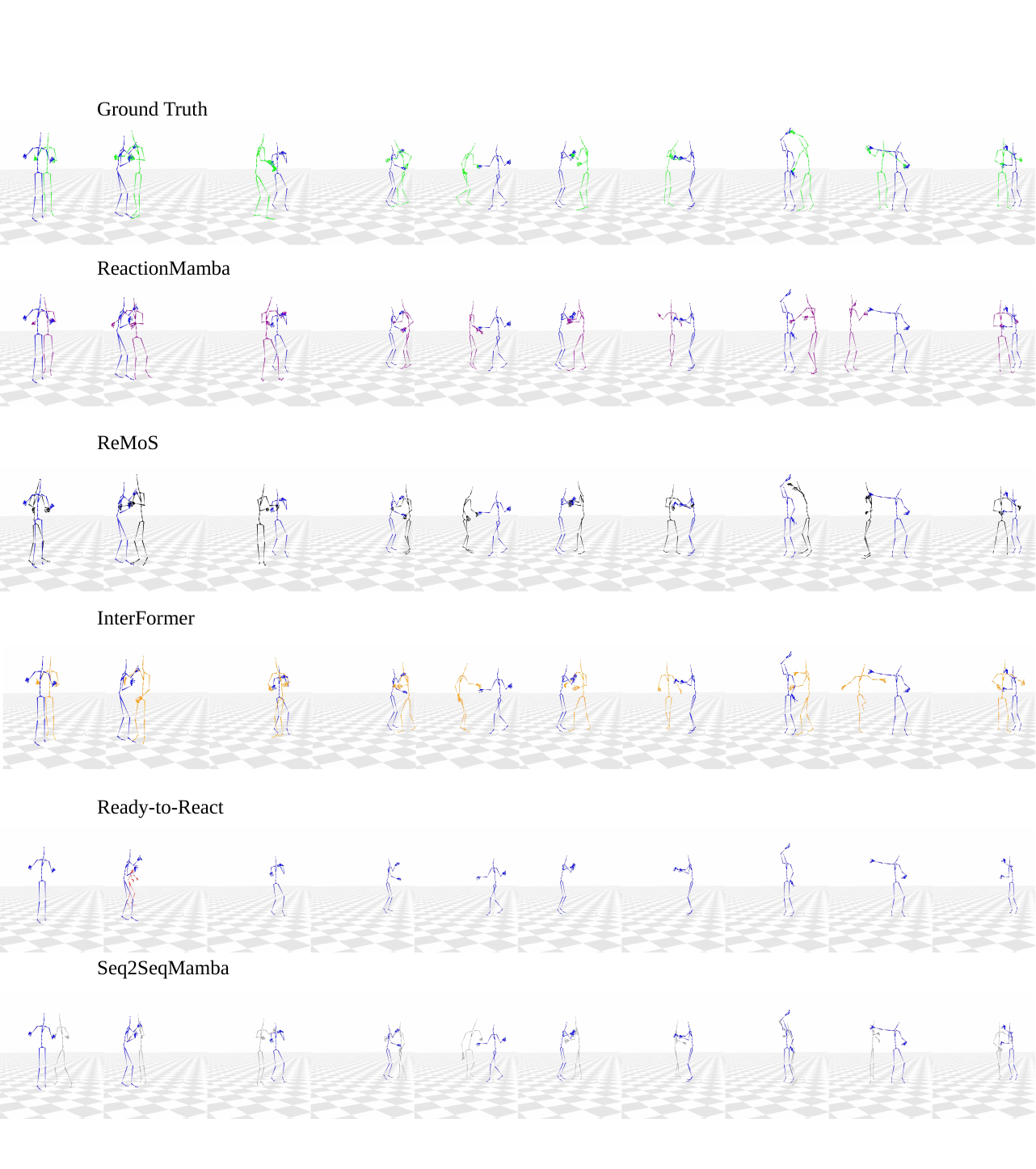}
    \vspace{0.1cm}
    \caption{ Visualization of sequence generated on \textbf{Lindy Hop dance}. In \textcolor{blue}{blue} the action motion used as condition. In \textcolor{green}{green} the ground truth reaction and in other colors the reaction generated by the different models. See the supplementary material for their corresponding animations}
    \label{fig:lh_long_pdf}
\end{figure*}

We illustrate long-term retrogressive generation on LindyHop (1000 frames) in Figure~\ref{fig:lh_long_pdf}. We observe that the qualitative results align well with the quantitative findings. The motion generated by ReactionMamba closely matches the ground truth and is comparable to Remos, while being hundreds of times faster. Interformer and the Seq2SeqMamba fails to produce meaningful hand interactions. Similarly, Ready-to-React diverges after the first few hundred frames, with the reactor continuing its dance-like motion far away from the actor, and the generated hand motions becoming incoherent and poorly articulated. Meanwhile, Overall, these results clearly demonstrate the better quality, stability, and efficiency of ReactionMamba compared to existing methods.


\end{document}